\documentclass{article}
\usepackage{ijcai23}

\usepackage{times}
\usepackage{soul}
\usepackage{url}
\usepackage[hidelinks]{hyperref}
\usepackage[utf8]{inputenc}
\usepackage[small]{caption}
\usepackage{graphicx}
\usepackage{amsmath}
\usepackage{amsthm}
\usepackage{booktabs}
\usepackage{algorithm}
\usepackage{algorithmic}
\usepackage[switch]{lineno}

\usepackage{gensymb}
\usepackage{textcomp}
\usepackage{amsfonts}
\usepackage{multirow}

\newcommand{\ie}{\textit{i}.\textit{e}.}
\newcommand{\eg}{\textit{e}.\textit{g}.}
\newcommand\mypara[1]{\noindent\textbf{#1}}

\newcommand{\etc}{\textit{etc}.}

\title{XFormer: Fast and Accurate Monocular 3D Body Capture}

\author{
Lihui Qian$^1$\thanks{X. Han and L. Qian contribute equally. X. Han is the corresponding author.}
\and
Xintong Han$^{1*}$
\and
Faqiang Wang$^{1}$
\and
Hongyu Liu$^{2}$
\and
Haoye Dong$^{3}$
\and \\
Zhiwen Li$^{1}$
\and
Huawei Wei$^{4}$
\and
Zhe Lin$^{1}$
\and
Cheng-Bin Jin$^1$
\affiliations
$^1$Huya Inc   \quad  $^2$Hong Kong University of Science and Technology\\
$^3$Carnegie Mellon University  \quad $^4$Tencent
\emails
turtleduck1995@gmail.com, hanxintong@huya.com
}

\begin{document}

\maketitle

\begin{abstract}
We present XFormer, a novel human mesh and motion capture method that achieves real-time performance on consumer CPUs given only monocular images as input. The proposed network architecture contains two branches: a keypoint branch that estimates 3D human mesh vertices given 2D keypoints, and an image branch that makes predictions directly from the RGB image features. At the core of our method is a cross-modal transformer block that allows information to flow across these two branches by modeling the attention between 2D keypoint coordinates and image spatial features. Our architecture is smartly designed, which enables us to train on various types of datasets including images with 2D/3D annotations, images with 3D pseudo labels, and motion capture datasets that do not have associated images. This effectively improves the accuracy and generalization ability of our system. Built on a lightweight backbone (MobileNetV3), our method runs blazing fast (over 30fps on a single CPU core) and still yields competitive accuracy. Furthermore, with an HRNet backbone, XFormer delivers state-of-the-art performance on Huamn3.6 and 3DPW datasets.
\end{abstract}

\section{Introduction}
\label{sec:intro}

3D body capture techniques play an essential role in a wide range of computer vision and graphics applications such as telepresence, VR chat, virtual YouTubers, and interactive gaming. However, accurate and temporally coherent body capture may require special (and usually expensive) devices, such as motion capture suits, multi-camera systems, and depth sensors, which highly hinder its large-scale applications. To resolve this issue, researchers have developed various methods that predict 3D body pose and mesh from monocular RGB images \cite{VNect_SIGGRAPH2017,kanazawa2018endtoend,kocabas2019vibe,mehta2020xnect,sun2021monocular,lin2020end,kocabas2021pare}. Despite remarkable progress has been made, these methods often fail to capture accurate body motion in challenging in-the-wild scenes, especially when real-time performance is desired.

One main challenge that remains in monocular 3D body capture is that acquiring training images with accurate 3D body annotations is hard. As a result, researchers attempt to use images with 2D annotations to facilitate the training. For example, most CNN-based approaches  \cite{kanazawa2018endtoend,kolotouros2019learning,mehta2020xnect,kanazawa2019learning,lin2020end} leverage datasets annotated with 2D keypoints (\eg, COCO \cite{lin2014microsoft}, LSP \cite{johnson2010clustered}, MPII \cite{andriluka_mpii2d_cvpr14}) and minimize the reprojection loss of keypoints in order to improve the accuracy on in-the-wild images. \cite{kocabas2019vibe,kanazawa2019learning} further extend the training modality to monocular videos with ground truth or pseudo 2D keypoint labels (PennAction \cite{zhang2013actemes}, PoseTrack \cite{andriluka2018posetrack}, InstaVariety \cite{kanazawa2019learning}) to exploit temporal information for boosting 3D motion estimation. Another line of research \cite{choi2020pose2mesh,martinez2017simple,zhao2019semantic} has shown that, without directly using image information, 2D keypoints alone provide essential (and sufficiently good) geometric information of 3D human pose/shape (\eg, short 2D hip joint distance suggests skinny lower body). And directly regressing 3D joints/mesh vertices from 2D keypoints is more effective and easier than previously thought. 

\begin{figure*}[t]
  \includegraphics[width=\textwidth]{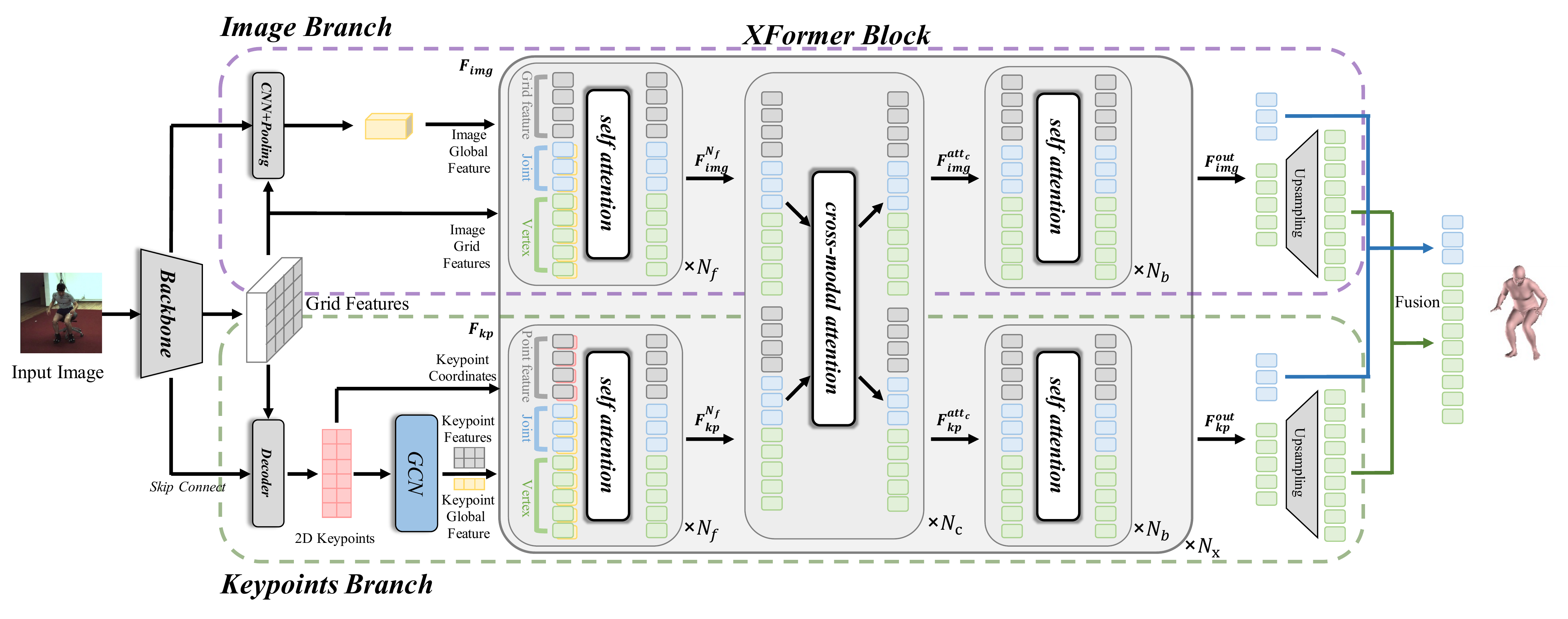}
  \vspace{-20pt}
  \caption{\textbf{XFormer system overview.} We predict 3D body joints and mesh with a network consisting of an image branch and a keypoint branch. These two branches interact with $N_x$ XFormer blocks, with each block containing $N_f$ front self-attention modules, $N_c$ cross-modal attention modules, and $N_b$ back self-attention modules. $F_{img}$ and $F_{kp}$ denote the image branch feature and the keypoint branch feature.
  }
  \vspace{-10pt}
  \label{fig:overview}
\end{figure*}

Intuitively, the semantic feature from images and 2D keypoints are complementary to each other, and it is interesting to integrate these two representations for a better pose and mesh reconstruction.
This has been investigated by concatenating 2D keypoint heatmaps with intermediate image features \cite{VNect_SIGGRAPH2017,mehta2020xnect}, adopting a trainable fusion scheme \cite{tekin2017learning}, and using a bilinear transformation for multi-modality fusion \cite{sun2019human}. However, these methods simply combine the multi-model features without explicitly exploiting their interactions. Meanwhile, these fusion strategies make the models unable to be trained on the Motion Capture data without paired images \cite{mahmood2019amass}, resulting in inadequate training.

To overcome the above limitations, we propose a real-time 3D body capture system, termed XFormer. The XFormer enforces knowledge transfer across image features and keypoints features with a proposed novel cross-modal attention module inspired by \cite{vaswani2017attention,lu2019vilbert}. Specifically, as shown in Figure \ref{fig:overview}, XFormer consists of two branches, \ie, the keypoint branch and the image branch. The keypoint branch takes the 2D keypoints to regress 3D joints and mesh vertices of the SMPL model \cite{SMPL:2015}, in which the 2D keypoints are predicted from the image by a keypoint detector or projected by the 3D keypoints from the MoCap data.
The image branch directly predicts the same information from the input image feature. In order to effectively integrate the advantages of these two branches, we feed the keypoint representations together with the image feature map into the proposed cross-modal attention module. By exchanging key-value pairs in the multi-head attention, we obtain the cross attention between the 2D keypoints and the image features, which enforces information communication between these two modalities. Extensive experiments in Section \ref{sec:ablation} demonstrate that this framework significantly outperforms each individual branch that only captures a single modality.

XFormer also takes full advantage of the datasets with different supervision types. In the previous work like \cite{kocabas2019vibe,kolotouros2021probabilistic,rempe2021humor}, the MoCap data (\eg, AMASS \cite{mahmood2019amass}), as it does not have paired images, is only used in the discriminator or as human motion priors. Thus, the strong supervision of 2D-3D pose mapping is ignored.
In contrast, our XFormer enjoys a ``modality plug-in'' characteristic, enabling the network to learn from MoCap or synthetic data without corresponding images. This is achieved by a modality switch, which uses an MLP to mimic the features learned by cross-modal attention.
Therefore, 2D keypoint features can skip the cross-modal attention and directly forward to the following network during training when the image modality is unavailable. Benefiting from this design, we can train on massive MoCap sequences even if they do not consist of any images.

With the proposed cross-modal attention and leveraging MoCap datasets, Xformer significantly boosts the performance when the backbone is lightweight (MobileNetV3 \cite{howard2019searching}) and achieves  real-time performance on consumer CPUs. In contrast, the accuracy of previous methods \cite{kocabas2019vibe,lin2021-mesh-graphormer,kocabas2021pare} largely drops by replacing their original heavy backbones (\eg, ResNet50 \cite{he2016deep}, HRNet \cite{wang2018high}) with lighter ones.

In sum, the main contributions of this paper are as follows:

1) We introduce a novel network architecture that estimates body joints and mesh from 2D keypoints and RGB image features, whose interactions are captured with a proposed cross-modal attention module.

2) The proposed two-branch architecture and XFormer blocks are designed to leverage all types of training data (2D and 3D, with and without images) of different modalities, which further improves its accuracy.

3) Our system with a light backbone takes less than 7ms per frame for an input person on an Nvidia GTX 1660 GPU and 30ms with a single thread of Intel i7-8700 CPU, obtaining significant speedup while maintaining satisfactory accuracy. 

4) Our proposed method achieves state-of-the-art quantitative performance with HRNet backbone on 3D pose estimation benchmarks and demonstrates significant qualitative results in challenging in-the-wild scenes.

% ----------------------------------------------------------

\section{Related Work}
\label{sec:related}

\mypara{3D Human Pose Estimation.}
3D human pose estimation can be categorized into model-free and model-based approaches.
Model-free approaches predict 3D human pose by directly estimating 3D keypoint from images \cite{pavlakos2018ordinal,lin2020end} or detected 2D human pose \cite{choi2020pose2mesh}.
Model-based methods estimate 3D pose by predicting body model (\eg, SMPL) parameters \cite{kanazawa2018endtoend,kocabas2019vibe,zanfir2021thundr} or meshes \cite{lin2020end,lin2021-mesh-graphormer}.
Our method directly predicts 3D mesh vertices of the SMPL model to leverage the strong non-local correlations among vertices.
Few prior approaches capture 3D human pose in real-time (\eg, \cite{VNect_SIGGRAPH2017,sun2021monocular}), as most methods use computationally demanding deep neural networks or/and require time-consuming kinematic optimization for post-processing.
Our method adopts a lightweight backbone with XFormer block heads, which runs much faster and occupies less memory, making it possible to run in real-time on even mobile devices while still performing on par with state-of-the-art methods.

\mypara{Human Pose Datasets.}
\label{sec:humanpose_datasets}
3D pose datasets are often created in controlled environments \cite{ionescu2013human3} or in a relatively small scale \cite{mehta2017monocular,von2018recovering}. MoCap datasets \cite{mahmood2019amass,mixamo}, on the other hand, provide massive 3D human motion sequences, but no corresponding images are available. Additionally, there are several datasets with pseudo 3D shape and pose labels. 
Intuitively, using all available datasets of different modalities (images/videos with annotated 2D/3D keypoints, fitted 3D models on unlabeled images, MoCap data without images, \etc) could improve performance. To allow the model to train on images with only 2D annotations, previous approaches commonly optimize the reprojection loss of keypoints \cite{kolotouros2019learning,mehta2020xnect}. 
VIBE further takes advantage of 2D video datasets for training a temporal model and uses MoCap data in the discriminator to force the output pose more temporally coherent. MoCap data is also used to build human motion priors for producing plausible and accurate motions \cite{rempe2021humor,kolotouros2021probabilistic}.
Similar to these methods, our method also aims to adopt data from different modalities. Nevertheless, we directly involve the MoCap data instead of just in a discriminator or prior, therefore fully utilizing the 3D information incorporated in this data.

\mypara{Transformers.}
Our cross-modal attention is built on transformers \cite{vaswani2017attention}. 
In the context of 3D pose modeling, transformer-based models are used to lift 2D keypoints to 3D \cite{li2021lifting,zhao2021graformer,li2022mhformer,shan2022p}, jointly model vertex-vertex and vertex-joint interactions \cite{lin2020end,lin2021-mesh-graphormer}, and focus on image regions that are relevant to the pose estimation \cite{zanfir2021thundr}. 
These methods usually adopt heavy backbones, and we empirically find that the performance significantly drops with a lightweight backbone or fewer transformer layers. 
In contrast, we use transformers from a different perspective in that information is transferred across keypoints and image features with cross-modal attention. This enables our model to maintain good performance with a lightweight backbone and a single-layer transformer encoder.

% ----------------------------------------------------------
\section{Method}
Figure \ref{fig:overview} summarizes our system. A lightweight feature extraction backbone (Section \ref{sec:feature}) takes the image as input, followed by an image branch (Section \ref{sec:feature}) and a keypoint branch (Section \ref{sec:keypoint}), both of which predict 3D joints and mesh vertices. 
The two branches interact with the XFormer blocks structure (Section \ref{sec:transformer}, Section \ref{sec:modalityswitch}) to exchange information between the keypoint modality and the image modality.
Finally, the outputs of two branches can be fused to bring about more precise and stable results (Section \ref{sec:ensemble}).

\subsection{Feature Extraction and Image Branch}
\label{sec:feature}
As shown in Figure \ref{fig:overview}, we first feed the person image $I \in \mathbb{R}^{H \times W \times 3}$ into a CNN to get the grid features and a pooled image global feature. Following \cite{lin2020end,lin2021-mesh-graphormer}, we tokenize these features together with the 3D coordinates of each mesh vertex and body joint of a coarse template mesh for positional encoding to obtain the image feature $F_{img}$. $F_{img}$ is then input to the image branch to recover 3D body mesh.

\subsection{Keypoint Branch}
\label{sec:keypoint}

The grid features extracted from the backbone are shared by the keypoint branch. We then adopt a heatmap-based method to estimate the 2D human pose. 
Following the common practice \cite{papandreou2017towards,bazarevsky2020blazepose}, a keypoint decoder inputs the grid features and low-level features from the backbone to predict keypoint heatmaps 
and offset maps 
of all 2D body keypoints, where $K$ is the number of body joints. Each heatmap $\mathcal{H}_k$ represents the confidence map of the corresponding keypoint, and the offset map $\mathcal{O}_k$ represents the offset ranging in $[-2, 2]$ to compensate for the lost accuracy since the heatmap width and height are a quarter of the resolution of the input image. The final predicted keypoint coordinates $ C \in \mathbb{R}^{K \times 2} $ are calculated by summing the coordinates of the maximum response value in the heatmap and the corresponding offset map value.
We then regress the 2D pose to obtain keypoint features using GCNs \cite{zhao2019semantic}. 
The final output of the GCN contains $K$ keypoint features, which are then concatenated with their corresponding 2D coordinates. We further use a mean-pooling layer to get the keypoint global feature and concatenate it with vertices and joints from the template mesh as in the image branch. Combining the aforementioned features, we obtain the keypoint branch feature $F_{kp}$ before inputting to the XFormer block.

\subsection{XFormer Block}
\label{sec:transformer}

\begin{figure}[t]
\centering
  \includegraphics[width=0.7\linewidth]{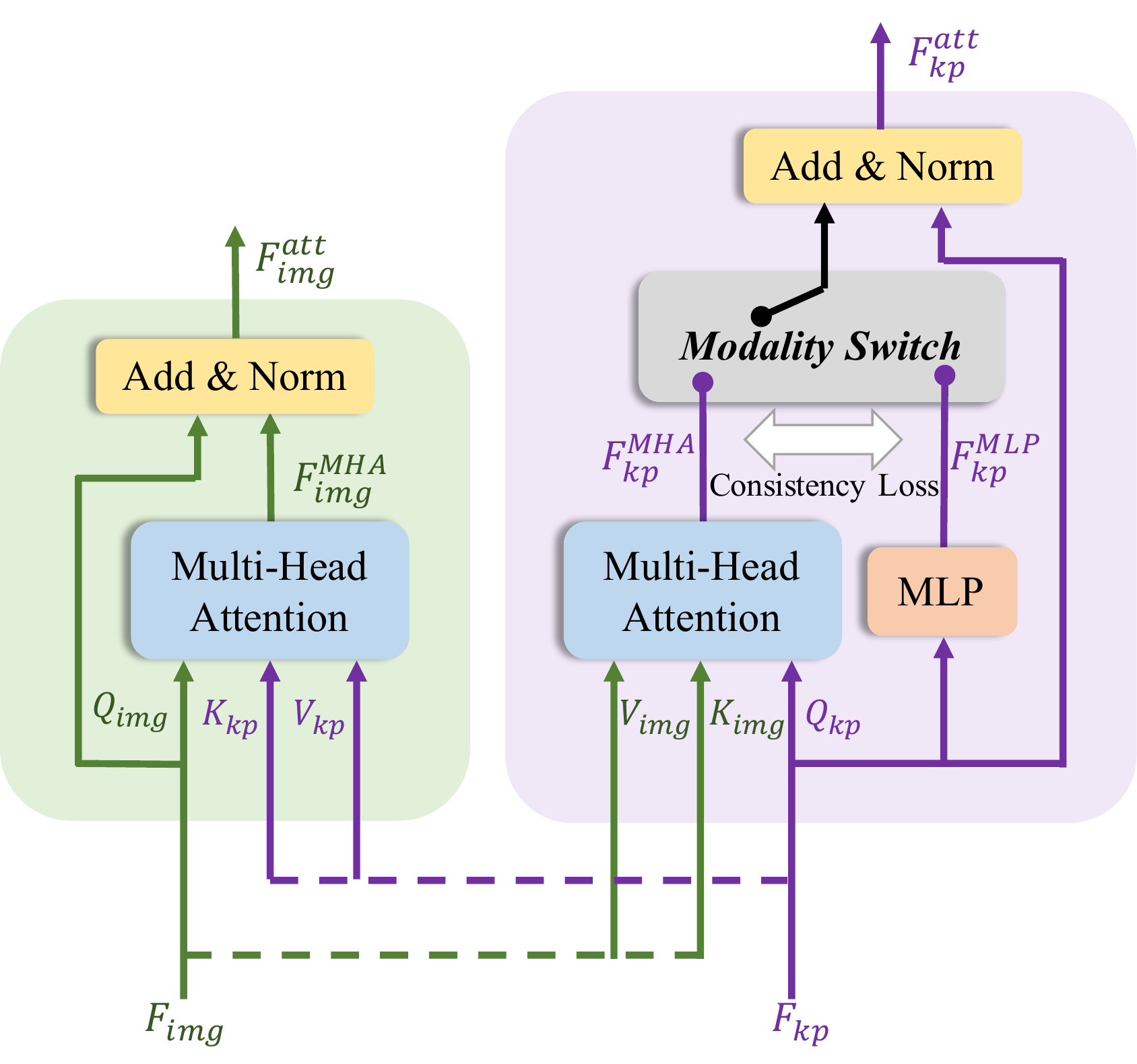}
  \caption{\textbf{Architecture of our proposed cross-modal attention module.} 
  We exchange the key-value pairs in the multi-head attention block of two modalities to model the interactions between image spatial features and 2D keypoints. Our cross-modal attention has a ``modality plug-in'' characteristic such that we can still train our network when the image modality is missing. When the image data is unavailable, we can switch the cross-modal feature $F_{kp}^{MHA}$ to the simulated feature $F_{kp}^{MLP}$ (see the ``modality switch'' block) without breaking the training data flow. 
  }
   \vspace{-10pt}
  \label{fig:transformer}
\end{figure}

We propose a model based on transformers \cite{vaswani2017attention} to encourage the information exchange between keypoints and image branches. Unlike previous transformer-based models \cite{lin2021-mesh-graphormer,lin2020end,li2021lifting,li2022mhformer}, we explicitly capture the attention between each 2D keypoint and the feature of each image location. Moreover, these two modalities are extracted from a shared backbone. Therefore, XFormer can be regarded as a cross-attention mechanism (for multiple modalities) and a self-attention mechanism (for the input image).

An XFormer block contains two types of attention modules, \textit{self-attention} modules, and \textit{cross-modal attention} modules. The self-attention module is a vanilla transformer encoder with multi-head attention that extracts the self-attended features of each branch. 
The structure of our cross-modal attention module is illustrated in Figure \ref{fig:transformer}. 
Specifically, the image branch feature $F_{img}$ that represents the visual features of the corresponding image spatial locations, is taken as one input to the cross-modal attention module. The other input is the keypoint branch feature $F_{kp}$ of 2D keypoints. 
We first obtain the query, key, and value matrices for each modality (\ie, $Q_{img}$, $K_{img}$, $ V_{img}$ of the image modality and $Q_{kp}$, $K_{kp}$, $ V_{kp}$ of the keypoint modality). Then, we exchange key-value pairs in the multi-head attention block of two modalities to get the feature $F_{img}^{MHA}$ and $F_{kp}^{MHA}$:
\begin{equation}
\label{eq:att_img}
\small 
F_{img}^{MHA} = \text{softmax}\left(Q_{img}K_{kp}^T / \sqrt{C_t}\right)V_{kp}
\text{,}
\end{equation}
\begin{equation}
\label{eq:att_kp}
\small
F_{kp}^{MHA} = \text{softmax}\left(Q_{kp}K_{img}^T / \sqrt{C_t}\right)V_{img}
\text{,}
\end{equation}
where $C_t$ is a scaling factor \cite{vaswani2017attention}. 
With the proposed XFormer block, the cross-modal attention matrix in the image branch provides rich spatial information to guide the network to focus on relevant regions given the keypoint coordinates. Meanwhile, the cross-modal attention matrix in the keypoint branch provides depth and semantic cues embedded in the image that helps regress better human body mesh. 

\subsection{Modality Switch}
\label{sec:modalityswitch}
To enable training with MoCap data that does not have the image modality, we design a novel 
modality switch
mechanism shown in Figure \ref{fig:transformer}. More specifically, we first input the feature $F_{kp}$ to an MLP ($F_{kp}^{MLP} = \text{MLP}(F_{kp})$). When image data is available, we apply a consistency loss (Eqn. (\ref{eqn:consistency-loss})) between $F_{kp}^{MLP}$ and $F_{kp}^{MHA}$ to supervise the MLP simulating the cross-modal attention. When training with MoCap data without images in the keypoint branch, we switch off the cross-modal attention and only train the MLP layer. Thus, the final attended features $F_{img}^{att}$ and $F_{kp}^{att}$ can be written as:
\begin{equation}
\label{eq:att_img_2}
\small 
F^{att}_{img} = \text{LN}\left(F_{img}^{MHA} + F_{img}\right)
\text{,}
\end{equation}
\begin{equation}
\label{eq:att_kp_2}
\small
F^{att}_{kp} = 
\begin{cases}
\text{LN}\left(F_{kp}^{MHA} + F_{kp}\right)\text{,}& F_{img}\text{ is available,} \\
\text{LN}\left(F_{kp}^{MLP} + F_{kp}\right)\text{,}& \text{otherwise,}
\end{cases}
% \text{,}
\end{equation}
where LN denotes the layer normalization used in the transformer.
As a result, the Xformer block does not rely on the presence of both modalities for training. We can drop the image branch and use projected 2D keypoints as input to the keypoint branch to train on MoCap data without paired images, while training on such data is impracticable for existing multi-modal 3D pose estimation approaches like \cite{VNect_SIGGRAPH2017,mehta2020xnect,tekin2017learning,sun2019human}. 

\subsection{Final Ensemble Result}
\label{sec:ensemble}
Finally, $F^{att}_{kp}$ and $F^{att}_{img}$ are further passed through self-attention modules to predict human mesh vertices and joints, as shown in Figure \ref{fig:overview}. 
The two branches both predict 3D joint locations, a coarse 3D body mesh with 431 vertices, and weak-perspective camera parameters. Similar to \cite{lin2020end,lin2021-mesh-graphormer}, we upsample the coarse mesh with MLPs to obtain the full SMPL mesh with 6,890 vertices. The estimations of two branches are fused to produce the final joints $J^{3D}$ and vertices $V^{3D}$: $J^{3D} = \lambda J^{3D}_{kp} + (1 - \lambda) J^{3D}_{img}$, $V^{3D} = \lambda V^{3D}_{kp} + (1 - \lambda) V^{3D}_{img}$, where $\lambda$ is simply set to $0.5$.

\section{Training}

\subsection{Loss Functions}
\label{sec:loss-function}
Our network is end-to-end trained by minimizing a total loss $L_{total}$ consisting of 2D keypoint detector loss $L_{map}$, keypoint branch loss $L_{kp}$, image branch loss $L_{img}$, and the consistency loss $L_{cons}$.

\begin{figure*}[t]
\centering
  \includegraphics[width=0.9\textwidth]{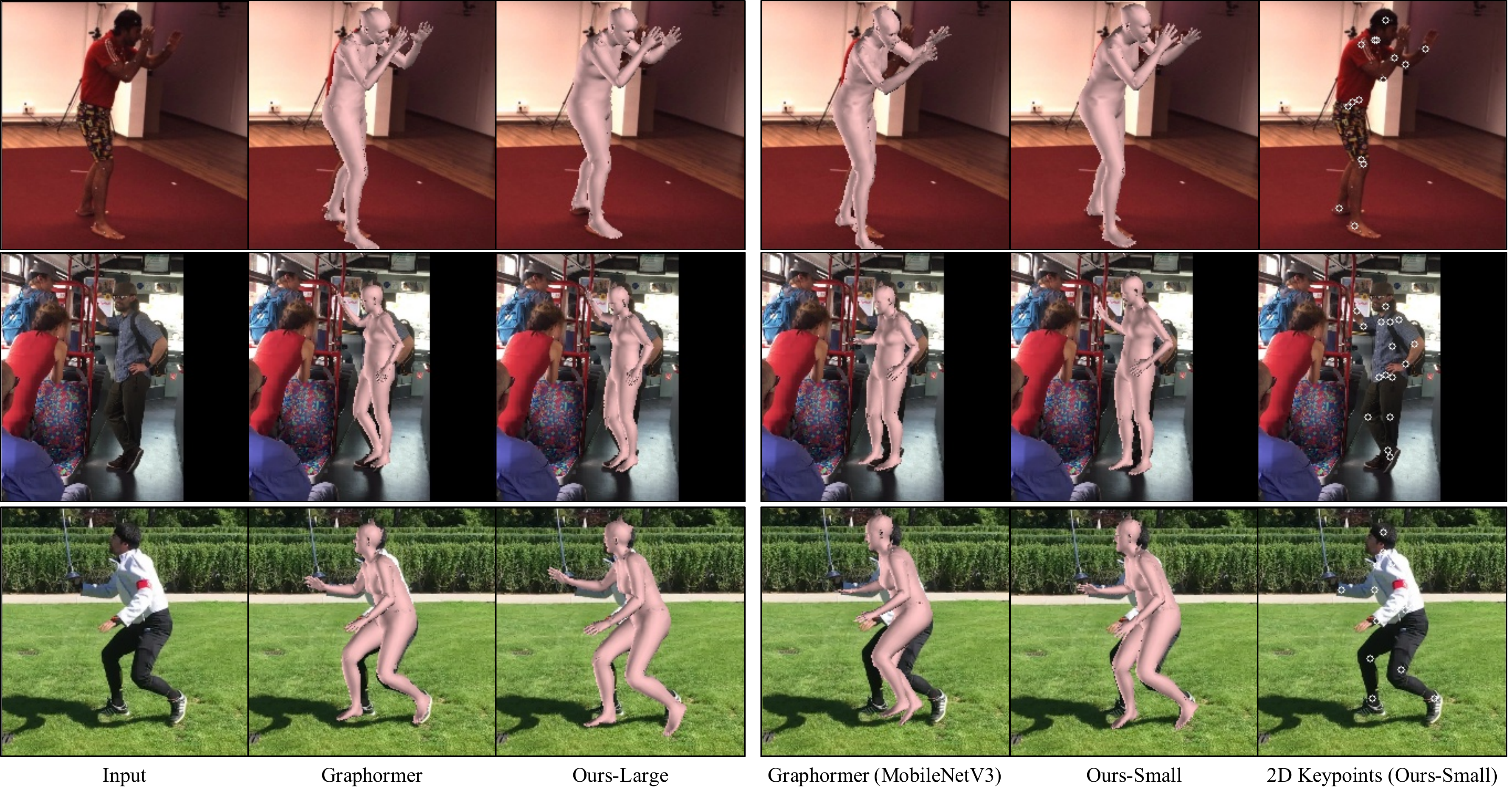}
  \vspace{-5pt}
  \caption{\textbf{Visual comparison} of our method against the previous state-of-the-art method Graphormer. XFormer performs slightly better than Graphormer  with a large backbone. While with a small backbone, XFormer reconstructs human mesh more accurately than Graphormer. The 2D keypoints predicted by our keypoint decoder are also visualized.}
  \label{fig:compair}
  \vspace{-10pt}
\end{figure*}

\mypara{2D Keypoint Detector Loss.}
We use a heatmap-based method to predict 2D keypoints in our keypoint branch. For the $k$-th keypoint, we create its ground-truth heatmap $\mathcal{\bar{H}}_k$ by a Gaussian distribution with mean as the keypoint coordinate and standard variation $\sigma = 2$. Each element of the ground-truth offset map $\mathcal{\bar{O}}_k$ is set to be the offset value w.r.t. the corresponding keypoint location when their distance is less than 2, otherwise, it is set to zero.
We minimize the L1 distance of the ground-truth and prediction:
$
    L_{map} =  \frac{1}{K}\sum_{i=1}^K w_k\left|\left|\mathcal{\bar{H}}_k - \mathcal{H}_k \right|\right|_1 + \frac{1}{K}\sum_{i=1}^K w_k\left|\left| \mathcal{\bar{O}}_k - \mathcal{O}_k \right| \right|_1
$
, where $w_k$ indicates the weight of each keypoint, and it is set to zero for the invisible keypoint.

\mypara{3D Reconstruction Loss.}
We optimize our framework by minimizing $L_{kp}$ for the keypoint branch and $L_{img}$ for the image branch. Specifically, we follow \cite{lin2020end} and formulate $L_{kp}$ as the sum of vertex loss ${L}^{V}_{kp}$, 3D joint loss ${L}^{J}_{kp}$, 3D joint regression loss ${L}^{{J}_{reg}}_{kp}$, and 2D re-projection ${L}^{{J}_{proj}}_{kp}$:
\begin{equation}
\small
\begin{aligned}
\label{eqn:vertex-loss}
{L}^{V}_{kp} = \frac{1}{M}\sum_{i=1}^{M} \left| \left| V^{3D}_{kp}-\bar{V}^{3D} \right| \right|_1,
\end{aligned}
\end{equation}
\begin{equation}
\small
\begin{aligned}
\label{eqn:3dpose-loss}
{L}^{J}_{kp} = \frac{1}{K}\sum_{i=1}^{K} \left| \left| J^{3D}_{kp}-\bar{J}^{3D} \right| \right|_1,
\end{aligned}
\end{equation}
\begin{equation}
\small
\begin{aligned}
\label{eqn:3dpose-reg-loss}
{L}^{{J}_{reg}}_{kp} = \frac{1}{K}\sum_{i=1}^{K} \left| \left| \mathcal W V^{3D}_{kp} - \bar{J}^{3D} \right| \right|_1,
\end{aligned}
\end{equation}
\begin{equation}
\small
\begin{aligned}
\label{eqn:2dpose-loss}
{L}^{{J}_{proj}}_{kp} = \frac{1}{K}\sum_{i=1}^{K} \left| \left| \Pi_{kp} J^{3D}_{kp} -\bar{J}^{2D} \right| \right|_1,
\end{aligned}
\end{equation}
where $\bar{V}^{3D}$, $\bar{J}^{3D}$, $\bar{J}^{2D}$ are the ground-truth 3D mesh vertex locations, the ground-truth 3D joint locations, and the ground-truth 2D keypoint coordinates. $M$ is the number of the vertices, $V^{3D}_{kp}$ denotes the output 3D vertex locations, and $J^{3D}_{kp}$ is the output 3D joint locations. With a pre-trained linear regressor $\mathcal W$, the 3D locations of body joints can be inferred from the 3D vertices by $\mathcal W V^{3D}_{kp}$ \cite{SMPL:2015}, and the 3D joint regression loss ${L}^{{J}_{reg}}_{kp}$ is their L1 distance to the ground-truth 3D locations. $\Pi_{kp}$ is the weak-perspective camera parameters predicted by the keypoint branch, which is used to obtain 2D projections of the 3D joints. The image branch loss $L_{img}$ is calculated in a similar fashion as $L_{kp}$.

\mypara{Consistency Loss.} As discussed in Section \ref{sec:transformer} and shown in Figure \ref{fig:transformer}, we apply a consistency loss $L_{cons}$ to make the MLP inside the cross-modal attention module simulate the cross-modal feature $F_{kp}^{MHA}$:
\begin{equation}
\begin{aligned}
\label{eqn:consistency-loss}
{L}_{cons} =  \left| \left| F_{kp}^{MHA}-F_{kp}^{MLP} \right| \right|_2.
\end{aligned}
\end{equation}
${L}_{cons}$ is only used in the keypoint branch when $F_{kp}^{MHA}$ is available (\ie, when training on the datasets with image data). 
This loss is conceptually similar to the idea of knowledge distillation \cite{hinton2015distilling}---we distill the knowledge in the cross-modal attention into an MLP layer to improve the model accuracy and robustness. 
But unlike knowledge distillation, we do not stop the gradient for $F^{MHA}_{kp}$, as the keypoint branch that learns from massive MoCap data and 3D labels provides valuable pose/shape priors, which enhances the performance of the image branch. 

\begin{table*}[t]
\centering
\resizebox{\textwidth}{!}
{
\begin{tabular}{lcccccccccccc}
\toprule
\multirow{2}{*}{Method} & \multirow{2}{*}{Backbone} & \multicolumn{4}{c}{Number of Blocks} && \multicolumn{2}{c}{Human3.6M}  && \multicolumn{3}{c}{3DPW}\\
\cline{3-6}\cline{8-9}\cline{11-13}
  &  & $N_{f}$ &$N_{c}$ & $N_{b}$&$N_{x}$ && MPJPE$\downarrow$  &  PA-MPJPE$\downarrow$ && MPJPE$\downarrow$  &  PA-MPJPE$\downarrow$ &  PVE$\downarrow$ \\ 
\midrule
HMR \cite{kanazawa2018endtoend}         
& ResNet50 & & & & &&    88.0      & 56.8      &&   130.0     & 76.7  &-   \\
SPIN \cite{kolotouros2019learning}      
& ResNet50 & & & & &&      -       & 41.1      &&  96.9      & 59.2  & 116.4   \\
VIBE \cite{kocabas2019vibe}             
& ResNet50 & & & & &&    65.6      & 41.4      &&   82.9      & 51.9  & 99.1   \\
XNect \cite{mehta2020xnect}            
& SelecSLS & & & & &&     63.6      & -         &&   134.2     & 80.3  &-    \\
ROMP \cite{sun2021monocular}            
& HRNet & & & & &&     -      & -         &&    76.7    &  47.3 &93.4      \\
Pose2Mesh \cite{choi2020pose2mesh}
& HRNet & & & & &&   64.9      & 47.0         &&   89.2     & 58.9  &-    \\
PARE* \cite{kocabas2021pare} 
& HRNet & & & & &&     54.4      & 38.2         &&   75.8    & 48.3  & 91.2    \\
METRO* \cite{lin2020end}             
& HRNet & 4 & 0 & 0 & 3 &&        54.0      & 36.7      &&   77.1       & 47.9   & 88.2  \\ 
Graphormer* \cite{lin2021-mesh-graphormer}             
& HRNet & 4 & 0 & 0 & 3 &&        53.3      & 36.1      &&   76.5 & 46.9  & 87.9   \\
Ours-Large w/o AMASS*                  
& HRNet & 1 & 1 & 2 & 3 &&        53.0      & 35.7      &&   75.8       & 46.0    &  87.5  \\
Ours-Large                  
& HRNet & 1 & 1 & 2 & 3 &&        \textbf{52.6}      & \textbf{35.2}      &&   \textbf{75.0}       & \textbf{45.7}    &  \textbf{87.1}  \\
\midrule
PARE* \cite{kocabas2021pare} 
& MobileNetV3 & & & & &&     72.8      & 46.6         &&   96.6    & 58.3  & 114.4    \\
Graphormer* \cite{lin2021-mesh-graphormer}             
& MobileNetV3 & 1 & 0 & 0 & 1 &&   74.6      & 50.1      &&    92.4      &   59.0 & 112.8   \\
Ours-Small w/o AMASS*           
& MobileNetV3 & 0 & 1 & 0 & 1 &&  71.1     & 45.8     &&    87.5      &     56.8 &  105.5 \\ 
Ours-Small             
& MobileNetV3 & 0 & 1 & 0 & 1 &&  \textbf{68.2}     & \textbf{44.2}     &&    \textbf{84.7}      &     \textbf{55.1} &  \textbf{102.6} \\ 
\bottomrule
\end{tabular}
}
\vspace{-5pt}
\caption{Performance comparison with state-of-the-art methods.``-'' denotes the results that are not available. Methods with ``*'' are trained with the same datasets (\ie, the first three types of datasets described in Section \ref{sec:train_dataset}) for a fair comparison. 
}
 \vspace{-10pt}
\label{tbl:compari_hm36}
\end{table*}

\subsection{Datasets}
\label{sec:train_dataset}
As discussed in Section \ref{sec:humanpose_datasets} and summarized in Table \ref{tbl:ablation_dataset}, common datasets can be divided into the following categories: 1) Image datasets with 3D annotations, such as 3DPW, UP-3D \cite{Lassner:UP:2017}, MuCo-3DHP \cite{mehta_3dv18}; 2) Image datasets with 2D keypoints annotations, such as COCO, MPII; 3) Image datasets with 2D keypoints annotations and pseudo 3D human labels, such as SPIN fits on COCO, Pose2Mesh fits on Human3.6M; 4) MoCap datasets without images, such as AMASS. Because of our system's ``plug-in'' characteristic, we can flexibly use all these kinds of datasets to train our network. 

Each training sample is used to minimize corresponding losses according to dataset type. Specifically, we use the 2D keypoint datasets COCO and MPII to train the network by minimizing $L_{map}$, $L_{kp}^{{J}_{proj}}$, $L_{img}^{{J}_{proj}}$ and $L_{cons}$. As for the image datasets with 3D annotations, they are used to train our whole network by minimizing the total loss $L_{total}$. 
For datasets without image data, we first generate 3D joint locations $\bar{J}^{3D}$ and 3D vertices locations $\bar{V}^{3D}$ from the ground-truth SMPL parameters. Then, we obtain 2D keypoints $\bar{J}^{2D}$ with a random orthographic projection. This way, we generate paired 2D keypoints input and ground-truth 3D joints and vertices output.
To improve the robustness of the keypoint branch, we employ the following data augmentations: 
1) we apply random rotations ($[-30\degree, 30\degree ]$, $[-30\degree, 30\degree]$, $[-60\degree, 60\degree]$ for row, pitch, and yaw, respectively) to the global rotation of the mesh to account for more projection variations; 
2) we apply random global shifting ($[-20, 20]$ pixels) and scaling ($[0.9, 1.1]$) to the 2D keypoints. 
As we fully utilize abundant data of different modalities across different domains, our modal can predict more plausible and accurate body mesh and joint locations. This is different from existing approaches that only use MoCap data to train a discriminator  \cite{kanazawa2018endtoend,kocabas2019vibe} or as shape and pose priors for regularization \cite{bogo2016keep}.

\section{Experiments}

\subsection{Main Results}
\label{sec:evaluation}

\mypara{Evaluation Protocol.}
\label{sec:Protocol}
We evaluate on the Human3.6M and 3DPW datasets following the protocols in \cite{kanazawa2018endtoend,kolotouros2019learning} and report Procrustes-aligned mean per joint position error (PA-MPJPE), mean per joint position error (MPJPE) and per-vertex error (PVE).

\mypara{Quantitative Evaluation.}
Table \ref{tbl:compari_hm36} compares our method with the prior works on Human3.6M and 3DPW. We compare these methods with small and large backbones. 
The original PARE trains with more powerful EFT-fitted \cite{joo2020exemplar} SMPL parameters on COCO, MPII, LSPET \cite{johnson_lspet_cvpr11} as pseudo 3D labels. To get a fair comparison, we train PARE with the same image datasets as Ours. 
As for the comparison with Graphormer, we use 3 full XFormer blocks in Ours-Large, which has similar network parameters. 
The results of Graphormer are reproduced with the official code released by the authors. 
We report the best performance of Graphormer and Xformer by running the experiments three times, and we find the results of Graphormer are more stochastic (on Human3.6M, Graphormer has a standard deviation of PA-MPJPE of 0.16, while that of Ours-Large is only 0.05). We attribute this to the fact that XFormer benefits from the stable complementary information provided by the keypoint branches, but Graphormer only relies on single modality input.
Note that we do not manage to reproduce the results of Graphormer reported in the paper (PA-MPJPE 34.5, MPJPE 51.2 on Human3.6M) in all three experiments.
For the methods with a small backbone (\eg, MobileNetV3), Ours-Small outperforms state-of-the-art methods Graphormer and PARE with the same backbone by a clear margin and running at a higher speed, as shown in Table \ref{tbl:compari_running_time}. 
For the methods with a large backbone (\eg, ResNet50, SelecSLS, and HRNet), Ours-Large performs better than the other methods on pose and shape estimation. 
Note that our model still performs favorably against state-of-the-art methods even if we turn off the modality switch to train without the MoCap dataset (\ie, Ours w/o AMASS), validating the effectiveness of our cross-modal attention. Powered with the ability of training with MoCap data, the performance of XFormer is further enhanced.

\mypara{Qualitative Evaluation.}
We conduct qualitative comparison against previous methods, as shown in Figure \ref{fig:compair}. These visual comparisons verify that our method outperforms previous real-time methods in 3D human mesh recovery and gives comparable results to state-of-the-art offline methods. 

\subsection{Ablation Study}
\label{sec:ablation}

\begin{table}[t]
\small
\centering

\resizebox{\columnwidth}{!}{%
\begin{tabular}{ll|l|l|l|l}
\toprule
\multicolumn{1}{l|}{Dataset Type}   & Datasets    & \multicolumn{4}{l}{PA-MPJPE}  \\ 
\midrule
\multicolumn{1}{l|}{\multirow{4}{*}{\begin{tabular}[c]{@{}l@{}}Image datasets\\
w/ 3D labels\end{tabular}}} & Human3.6M                 & \multirow{4}{*}{\checkmark} & \multirow{4}{*}{\checkmark} & \multirow{4}{*}{\checkmark}  & \multirow{4}{*}{\checkmark} \\
\multicolumn{1}{l|}{}   & UP-3D                &                            &                            &                            \\
\multicolumn{1}{l|}{} & MuCo-3DHP           &                            &                            &                            \\ 
\multicolumn{1}{l|}{} & 3DPW  &                            &                            &                            \\ 
\midrule
\multicolumn{1}{l|}{\multirow{2}{*}{\begin{tabular}[c]{@{}l@{}}2D keypoint datasets\\ 
w/o pseudo 3D labels\end{tabular}}}               & COCO                   & \multirow{2}{*}{\checkmark} & \multirow{2}{*}{\checkmark} & \multirow{2}{*}{\checkmark} & \multirow{2}{*}{\checkmark} \\
\multicolumn{1}{l|}{} & MPII           &                            &                            &                            \\ 
\midrule
\multicolumn{1}{l|}{\multirow{2}{*}{\begin{tabular}[c]{@{}l@{}}2D keypoint datasets\\ w/ pseudo 3D label\end{tabular}}}   & SPIN fits              & \multirow{2}{*}{} & \multirow{2}{*}{}          & \multirow{2}{*}{\checkmark}  & \multirow{2}{*}{\checkmark}\\
\multicolumn{1}{l|}{}                                                                                                     & Pose2Mesh fits         &                            &                            &                            \\ 
\midrule
\multicolumn{1}{l|}{\multirow{2}{*}{\begin{tabular}[c]{@{}l@{}}MoCap datasets\\ w/o images\end{tabular}}}                 & \multirow{2}{*}{AMASS} & \multirow{2}{*}{} & \multirow{2}{*}{\checkmark} & \multirow{2}{*}{}   & \multirow{2}{*}{\checkmark}       \\
\multicolumn{1}{l|}{}                                                                                                     &                        &                            &                            &                            \\ 
\midrule
\multicolumn{2}{l|}{Ours-Small}                                                                                                 &        47.0     &    46.5                         &                   45.8 &                   \textbf{44.2}      \\
\multicolumn{2}{l|}{Ours-Large}                                                                                                               &     36.9            &                 36.4           &  35.7  &  \textbf{35.2}    \\

\bottomrule
\end{tabular}
}
\caption{Ablation study on Human3.6M by varying the types of training datasets used in our method. With the proposed modality switch mechanism, we achieve performance improvement by taking advantage of different types of datasets.
}
\vspace{-5pt}
\label{tbl:ablation_dataset}
\end{table}

\mypara{On Different Training Datasets.} As the proposed XFormer can leverage datasets with different annotation types, we compare on different combinations of datasets in Table \ref{tbl:ablation_dataset} to evaluate the effect of different types of datasets. As the first two types of datasets are commonly used in all related works, we always turn them on. It shows the MoCap datasets without images can bring about evident performance gain as the modality switch mechanism enables XFormer to train on such data. Note that state-of-the-art methods all leverage 3D pseudo-labeled datasets, and Table \ref{tbl:ablation_dataset} also shows that such datasets benefit XFormer.
Plus, dropping both the 3D pseudo labeled and MoCap data, we still outperform the best models trained without these types of data \cite{kocabas2019vibe,mehta2020xnect}.

\begin{table}[t]
\renewcommand{\tabcolsep}{5pt}
\centering
{
\resizebox{\linewidth}{!}{

\begin{tabular}{lccccc}
\toprule
\multirow{2}{*}{Method}         & \multicolumn{2}{c}{Human3.6M} && \multicolumn{2}{c}{3DPW} \\ 
\cline{2-3}\cline{5-6}
& MPJPE & PA-MPJPE && MPJPE & PA-MPJPE \\
\midrule
Image Branch Only       &   74.5    & 51.3 && 93.3 & 60.2   \\
Keypoint Branch Only   &   72.8      & 50.5  && 94.5 & 61.6  \\  
w/o Consistency Loss &   70.3      & 45.1  && 86.6 & 55.8  \\ 
\midrule
Single Branch with Both Tokens &   71.4      & 49.2  && 91.2 & 58.5  \\ 
Ours-Small (w/o AMASS) &   71.1      & 45.8  && 87.5 & 56.8  \\ 
\midrule
Keypoint Output     &   68.8    & 45.0  && 85.9 & 56.3     \\
Image Output        &   69.0    & 45.1  && 86.8 & 57.0 \\
Ours-Small     &   \textbf{68.2}    & \textbf{44.2}  && \textbf{84.7} & \textbf{55.1}  \\
\bottomrule
\end{tabular}

}
}
 \vspace{-5pt}
\caption{Ablation study on different settings of our small model. 
Without our two-branch structure, the models trained with a single branch drop significantly. Our cross-modal attention module effectively models the information exchange of two branches. 
A single-branch model with both tokens does perform better than individual branches but is still worse than a two-branch model with cross-modal attention.
For a better understanding of the contribution of our cross-modal attention mechanism, we also put the results trained without the AMASS dataset. 
``w/o Consistency Loss'' means we remove the consistency loss.
Ours Keypoint Output and Image Output indicate that we do not use the ensemble strategy and inference with the full network (with cross-modal attention module) only with the output of the keypoint branch $V^{3D}_{kp}$ or the image branch $V^{3D}_{img}$, respectively. 
 \vspace{-20pt}
}
\label{tbl:compari_branch}
\end{table}

\mypara{On Two Branches and Cross-Modal Attention.}
To validate the effectiveness of our two-branch strategy and the cross-modal attention module, we test several single-branch models, including the model containing only an image branch (Image Branch Only), the model containing only a keypoint branch (Keypoint Branch Only), and the model which takes image and keypoint tokens sequentially as input (Single Branch with Both Tokens). The results are shown in Table \ref{tbl:compari_branch}.
Furthermore, the output of each branch from the full two-branch network still works better than the corresponding single-branch network when inferencing. 
We attribute these observations to two reasons: 1) the cross-modal attention module effectively models the information interactions between the keypoint and image modalities; 2) the additional MoCap datasets improve the generalization of the keypoints branch, resulting in improved full network performance.
The ensemble strategy further improves performance.

\mypara{On Consistency Loss.} 
As shown in Table \ref{tbl:compari_branch}, the performance drops if we remove the consistency loss and do not let the MLP mimic the cross-modal attention feature. 
The consistency loss can distill the knowledge from cross-modal attention into the MLP layer, which gains performance when training with both the image and MoCap data.

\mypara{Discussions.} We observe that the improvement of our XFormer is more notable for small backbones. We attribute this to that for heavy backbones, both branches have the network capacity of learning fairly good 3D body mesh, and adding cross-modal attention and making use of the MoCap dataset mildly improve the performance. 
As the model size decreases, small models have lower generalization ability and are more prone to appearance domain gap between limited controlled environment data \cite{mehta2017monocular,von2018recovering} and large-scale in-the-wild images, making it hard to predict accurate 3D body shape directly from image features. 2D keypoints, which are easier to estimate thanks to well-established datasets and methods  \cite{cao2019openpose,bazarevsky2020blazepose}, provide complementary information to boost the small model’s performance. This further validates that XFormer is suitable for light backbones in real-time scenarios while existing methods have severely degraded performance when the model capability decreases.

\subsection{Running Time Analysis}
\label{sec:runtime}

\begin{table}[t]
\small

\renewcommand{\tabcolsep}{1.0pt}
\centering
\resizebox{\linewidth}{!}{%
\begin{tabular}{lcccc}
\toprule
\multirow{2}{*}{Method}    & GPU Speed & CPU Speed \\ 
    & (fps) & (fps) \\ 
\midrule
VNect \cite{VNect_SIGGRAPH2017}        & 30.3 & 3.1 \\
SPIN \cite{kolotouros2019learning}       & 126.6 & 7.9 \\
VIBE \cite{kocabas2019vibe}             & 123.4 & 8.0    \\
XNect \cite{mehta2020xnect}            & 29.8 & 2.9  \\ 
ROMP \cite{sun2021monocular}   & 24.2 & 3.1 \\ 
PARE \cite{kocabas2021pare} (MobileNetV3) & 103.1 & 27.6 \\
Graphormer \cite{lin2021-mesh-graphormer} (MobileNetV3)    & 142.1 & 32.4    \\
Ours-Small (MobileNetV3)                        & \textbf{154.0} & \textbf{37.6}    \\
\bottomrule
\end{tabular}
}
\vspace{-5pt}

\caption{Inference time of state-of-the-art methods and XFormer. The CPU speeds of our methods are tested with a single thread, while VNect and XNect use multiple CPU cores.}
\vspace{-10pt}
\label{tbl:compari_running_time}
\end{table}

In Table \ref{tbl:compari_running_time}, we profile the running time of our method on a desktop with an Intel(R) Core(TM) i7-8700 CPU @ 3.20GHz and an Nvidia GeForce GTX 1660. All CPU models are accelerated with OpenVino.
Our network achieves real-time performance (154.0fps 
on GPU and 37.6fps on CPU). We observe that our method achieves a good balance between effectiveness and efficiency. 
Our method gives a comparable reconstruction error and runs much faster compared with most approaches (\eg, \cite{kolotouros2019learning,kocabas2019vibe}). We have a similar speed to Graphormer (MobileNetV3) while obtaining much more accurate estimations.

\section{Conclusion and Limitations}
In this paper, we have described a fast and accurate approach to capturing the 3D human body from monocular RGB images. We utilize all available datasets of different modalities by designing an effective two-branch network to predict 3D body joints and mesh jointly. The information incorporated in these two branches interacts through a novel cross-modal attention module. Experiments have demonstrated that our system runs at more than 30fps on consumer CPU cores while still achieving accurate motion capture performance.

However, XFormer faces several limitations: 1) Although fast, XFormer is a top-down approach, and the running time is nearly linear to the number of people in the image. 
2) XFormer does not explicitly exploit temporal information. We believe that modeling sequential data could improve the performance of video inference. 3) The monocular 3D human mesh recovery task is ill-posed, and XFormer still performs worse than state-of-the-art multi-view approaches. We believe that XFormer can be extended in a multi-view setting by not only modeling cross-modality attention in each view but also modeling cross-view attention with a similar transformer architecture.

\bibliographystyle{named}
\bibliography{XFormer_bib}

\appendix
% This supplementary material contains extra implementation details, visual results, ablation studies, and discussions.
% We will release our code to facilitate further research. Moreover, we will also make a more accurate (trained with in-house data) model and our C++ SDK/library public, which can be readily used by downstream real-time applications.

\begin{figure*}[]
  \centering
  \includegraphics[width=1\textwidth]{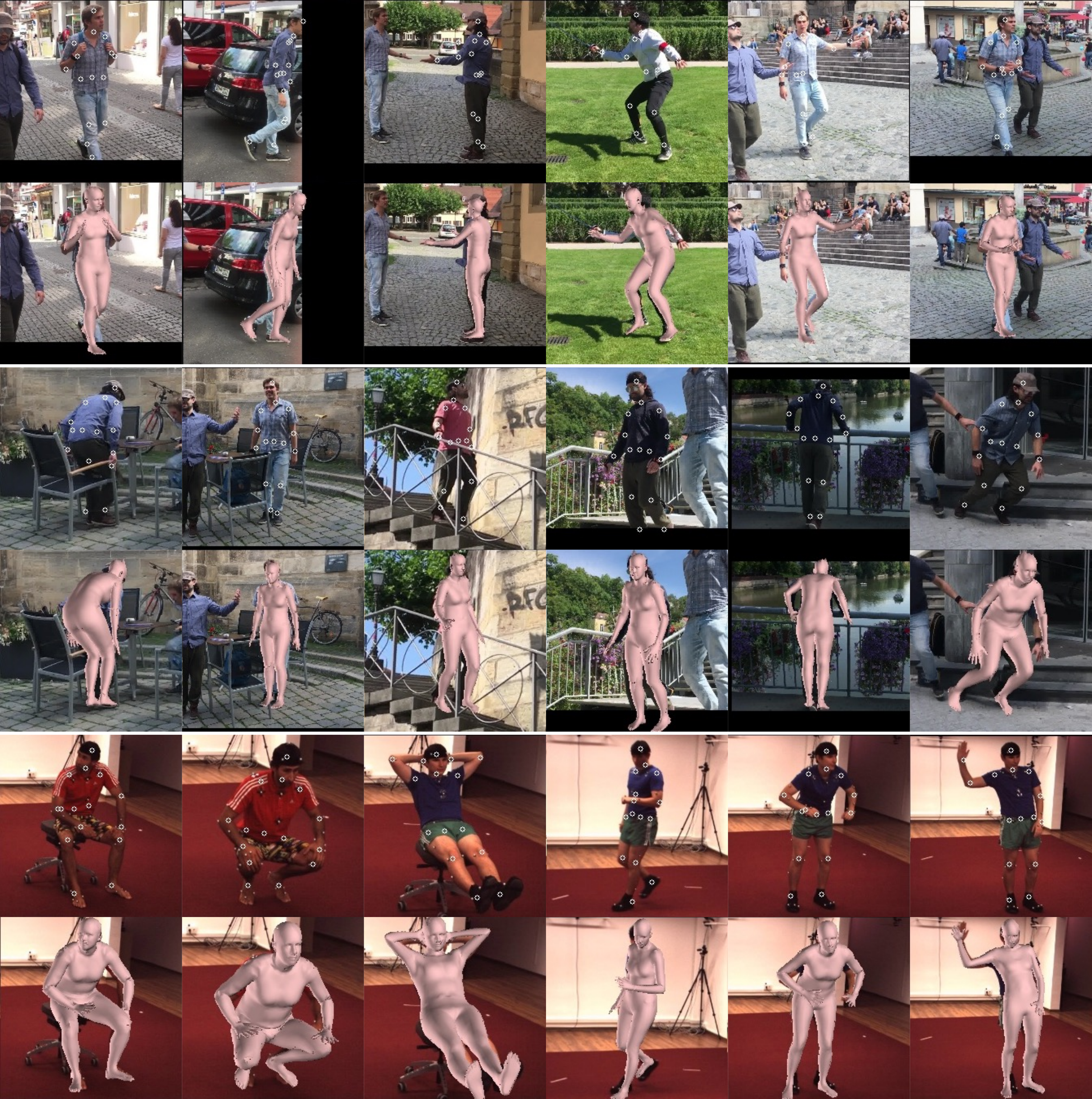}
  \caption{More results of Ours-Small model. The odd rows show the predicted 2D keypoints, and the even rows are the predicted body meshes.} 
  \label{fig:results}
\end{figure*}

\section{Implementation Details}

\mypara{Keypoint Decoder Architecture.} 
The keypoint decoder in the keypoint branch contains three transposed convolutional layers, which gradually increase the feature map resolution. These transposed convolutional layers receive not only the output feature of the previous layer, but also the skip connections from the backbone as in \cite{ronneberger2015u} to allow the information flow through network layers.

\mypara{Training.}
We adopt the Adam optimizer \cite{kingma2014adam}.
We use HRNet-W64, MobileNetV3 as the backbone in Ours-Large and Ours-Small, respectively.
We train the whole network for 200 epochs and select the best performing models on the validation set. The training process of the Ours-Small takes about 3 days on 4 NVIDIA V100 GPUs, and Ours-Large takes about 5 days on 8 NVIDIA V100 GPUs. When training PARE \cite{kocabas2021pare}  with the MobileNetV3 backbone, we find that it fails to achieve good convergence when training the network with ImageNet pre-trained weights (only 61.9 PA-MPJPE when evaluated on 3DPW). We only manage to get decent results (58.3 PA-MPJPE) after first pre-training the backbone on the 2D keypoint detection task with the COCO dataset followed by fine-tuning on 3D human body estimation. As opposed to PARE, Ours-Small does not rely on pre-training on COCO, which again verifies that XFormer enjoys higher generalization ability with small backbones.

\mypara{Detailed Evaluation Protocol}. We evaluate our system on the Human3.6M \cite{ionescu2013human3} and 3DPW \cite{von2018recovering} datasets. For Human3.6m, we train XFormer on 5 subjects (S1, S5, S6, S7, S8) and test on 2 subjects (S9, S11) following the protocol as in \cite{kanazawa2018endtoend,pavlakos_volumetric_cvpr17}. For 3DPW, 14 joints out of 17 estimated joints are evaluated as in \cite{kanazawa2018endtoend,pavlakos2018humanshape,kolotouros2019learning,kolotouros2019convolutional}.

\section{Additional Results}

\mypara{Evaluation on MPI-INF-3DHP.}
In addition to Human3.6M and 3DPW, we compares our method with PARE and Graphormer on MPI-INF-3DHP \cite{mehta2017monocular}, as shown in Table \ref{tab:3DHP}. Note that all these methods do not train on MPI-INF-3DHP training set (see Section 5.1). Our method outperforms the other methods with both the small backbone (\ie, MobileNetV3) and the large backbone (\ie, HRNet) by a clear margin.

\begin{table}[h]
\footnotesize
\renewcommand{\tabcolsep}{2pt}
\centering
\begin{tabular}{lcc}
\toprule
MPI-INF-3DHP & MPJPE & PA-MPJPE  \\ 
\midrule
PARE (\textit{HRNet})      &   111.7   &  67.5   \\
Graphormer (\textit{HRNet})      &   112.5   &  66.8   \\
Ours-Large (\textit{HRNet}) &   \textbf{109.8}   &  \textbf{64.5}  \\
\midrule
PARE (\textit{MobileNetV3})    &  130.2  &  76.6  \\  
Graphormer (\textit{MobileNetV3})    &   134.6   &  78.1   \\  
Ours-Small (\textit{MobileNetV3}) &   \textbf{127.3}   &  \textbf{73.2 } \\
\bottomrule
\end{tabular}
\caption{ Results on MPI-INF-3DHP dataset.}
\label{tab:3DHP}
\end{table}

\mypara{Shape Estimation.} 
In addition to report the accuracy of the estimated 3D poses, Table \ref{tbl:supp_ablation_consistency} shows the shape estimation performance, measured by PVE, of different settings. The XFormer blocks improve the PVE of the keypoint branch from 120.1 to 106.8 due to the complementary information from features of the image branch. And we find that incorporating the 3D pose/shape priors learned from 2D keypoints boosts the performance of the image branch (improving PVE from 118.4 to 108.1).

\begin{table}[h]
\renewcommand{\tabcolsep}{0.5pt}
\centering
\resizebox{\columnwidth}{!}{
\begin{tabular}{lccccccccccc}
\toprule
\multirow{2}{*}{Methods}         & \multicolumn{3}{c}{MPJPE} & \multirow{2}{*}{} & \multicolumn{3}{c}{PA-MPJPE} & \multirow{2}{*}{} & \multicolumn{3}{c}{PVE} \\ 
\cline{2-4}  \cline{6-8} \cline{10-12}
& Final & KB & IB && Final & KB & IB && Final & KB & IB \\
\midrule
Image Branch Only (MobileNetV3) &      -&  -& 93.3   &&     -&   -&  60.2   &&     -&  -& 118.4 \\
Keypoint Branch Only (MobileNetV3) &       -&  94.5 &-    &&    -& 61.6 &-    &&     -& 120.1  &-  \\
Ours-Small w/o Consistency Loss &    86.6  & 89.2 & 87.5   && 55.8  & 56.9 & 57.5   &&   105.0 & 106.8 & 108.1 \\
Ours-Small &                      84.7  & 85.9 & 86.8   && 55.1  & 56.3 & 57.0   &&   102.6 & 104.7 & 107.3 \\
\bottomrule
\end{tabular}
}
\caption{ Ablation study on consistency loss on 3DPW dataset. KB, IB denote Ours Keypoint Output and Ours Image Output (see Table 3 in the main paper), respectively.
}
\label{tbl:supp_ablation_consistency}
\end{table}

\mypara{Visual Results.}
Figure \ref{fig:results} visualizes more qualitative 3D body capture results of the Ours-Small model. Our proposed method is robust against clustered backgrounds and occlusions due to the compensation between our image branch and keypoint branch.

\begin{figure*}[]
  \centering
  \includegraphics[width=1\textwidth]{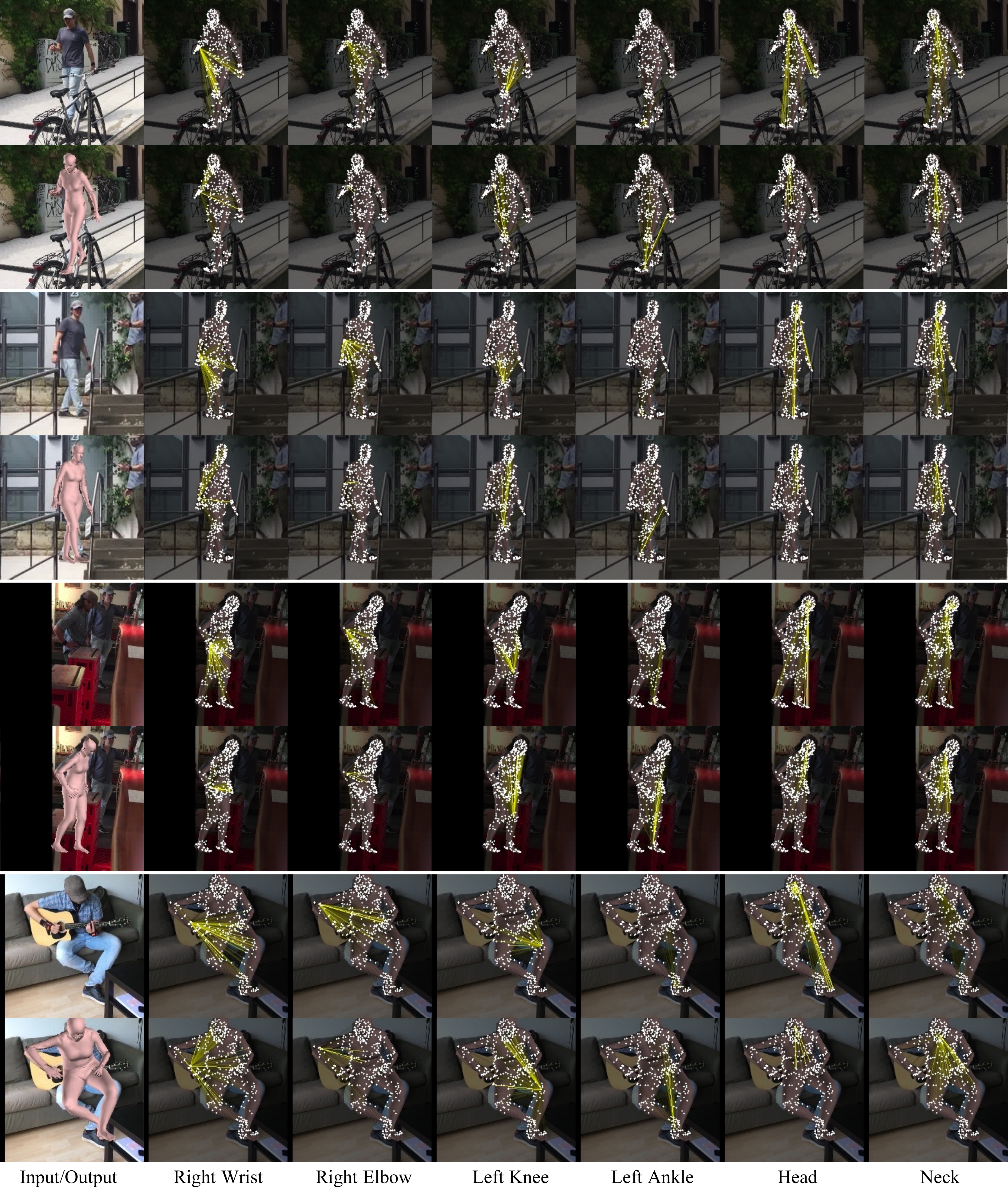}
  \caption{
  Visualization of the cross-modal attention matrix in two branches. The first column is the input and the output of the Ours-Small model. 
  The odd rows are the cross-modal attention matrix in the keypoint branch, and the even rows are the cross-modal attention matrix in the image branch.
  } 
  \label{fig:attention}
\end{figure*}

\mypara{Visualization of Cross-modal Attentions.} 
We visualize our cross-modal attentions in the keypoint branch and the image branch, respectively, in Figure \ref{fig:attention}. 
The cross-modal attention matrix tends to focus on different parts of the body, which is in line with our expectations that the semantic features from images and keypoints are complementary to each other. 
For example, as shown in the second column, when the model predicts the vertices of right wrist, the cross-modal attention in the keypoint branch focuses on the lower body vertices, while the cross-modal attention in the image branch focuses more on the left wrist and the right upper arm. The attention of the head joint in the keypoint branch focuses on the ankle vertices, while the attention in the image branch focuses on the upper body mesh.

Furthermore, the cross-modal attention matrix in the keypoint branch focuses on global body parts, even if the part is occluded. In contrast, the matrix in the image branch focuses on local and non-occluded body parts. 
We speculate that the adjacency matrix in GCN may help the keypoint branch to extract the global feature, while for the image branch, local features with valid semantics may help improve the accuracy of predicted vertices.

\begin{table}[t]
\small
\renewcommand{\tabcolsep}{5pt}
\centering
{
\resizebox{\linewidth}{!}{

\begin{tabular}{lccccc}
\toprule
\multirow{2}{*}{Method} &  \multirow{2}{*}{Backbone} & \multirow{2}{*}{Modality} & \multirow{2}{*}{Attention} & Modality      & Human3.6M \\ 
&&  &  & Switch &  PA-MPJPE  \\
\midrule
PARE      &   HRNet    & image  & part-attention &  &  38.2 \\
DSD      &   HRNet    & image, kp  & bilinear transformation &  & 44.3 \\
Lifting Transformer & CPN & kp & strided transformer  & & 36.1 \\
METRO    &   HRNet      & image & self-attention &   & 36.7\\  
Graphormer &   HRNet     & image & self-attention &  & 36.1 \\ 
Ours-Large w/o AMASS &  HRNet      & image, kp  & self, cross-attention &  &  35.7 \\ 
Ours-Large  &   HRNet      & image, kp & self, cross-attention & activated & \textbf{35.2} \\ 
\midrule
PARE   &   MBV3    & image & part attention &   & 46.6\\
Graphormer  &   MBV3    &image & self-attention &   & 50.1 \\
Ours-Small w/o AMASS        &   MBV3    & image, kp & self, cross-attention &   &  45.8\\
Ours-Small     &   MBV3    & image, kp & self, cross-attention & activated & \textbf{44.2} \\
\bottomrule
\end{tabular}
}
}
 \vspace{-5pt}
\caption{Comparison with transformer-based methods. 
}
 \vspace{-10pt}
\label{tbl:compari_transformer}
\end{table}

\mypara{Comparison of Attentions.} 
We compare the XFormer with other methods using attention in Table \ref{tbl:compari_transformer}. 
METRO and Graphormer use only image modality. Without the geometric information of keypoints, replacing the backbones of these methods with MBV3 leads to a substantial drop in accuracy (PA-MPJPE 36.1$\rightarrow$50.1 of Graphormer).
Lifting Transformer \cite{li2021lifting} use only keypoint modality. 
DSD \cite{sun2019human} uses two modalities, but it simply combines features by bilinear transformation without explicitly exploiting their interactions. 
Furthermore, the modality switch enables the XFormer to train on MoCap data.
These advantages contribute to the improved accuracy and robustness of the XFormer.

\section{Additional Ablation Study}

\mypara{On GCN.} 
Table \ref{tbl:supp_ablation_gcn} shows the comparison between different settings of the GCNs. 
We replace each GConv with a single-layer MLP, the PA-MPJPE becomes 2.2 mm worse. 
Moreover, concatenating the 2D keypoints coordinates in the keypoint representation $F_{kp}$ improves PA-MPJPE from 44.8 to 44.2. 

\begin{table}[h]
\small
\renewcommand{\tabcolsep}{5pt}
\centering
{
\begin{tabular}{lcc}
\toprule
Methods         & ~MPJPE~ &  ~PA-MPJPE~ \\ 
\midrule
Ours-Small (MLP)       &   72.7    & 46.4    \\
Ours-Small (w/o 2D Keypoint)    &   69.4      & 44.8     \\  
Ours-Small (GCN) &   68.2    & 44.2   \\
\bottomrule
\end{tabular}
}
\caption{Ablation study on different GCN settings, evaluated on the Human3.6M dataset. ``Ours-Small (MLP)" replaces the GConv with MLP.
``Ours-Small (w/o 2D Keypoint)" indicates that we do not concatenate the 2D keypoints in the input features. 
}
\label{tbl:supp_ablation_gcn}
\end{table}

\begin{table}[h]
\small
\renewcommand{\tabcolsep}{1pt}
\centering
{
\resizebox{\columnwidth}{!}{
\begin{tabular}{lcc}
\toprule
Methods         & ~MPJPE~ &  ~PA-MPJPE~ \\ 
\midrule
Image Branch Only (\textit{HRNet})    &   53.3      & 36.1     \\  
Ours-Large (w/o AMASS)  &   53.0    & 35.7   \\
Ours-Large  &   52.6    & 35.2   \\
\midrule
Add Fusion (\textit{MobileNetV3})              &   74.8      & 49.9     \\ 
Concat Fusion (\textit{MobileNetV3})           &   76.0      & 53.9     \\
Image Branch Only (\textit{MobileNetV3})     &   74.5      & 51.3     \\
Keypoint Branch Only (\textit{MobileNetV3})   &   72.8      & 50.5     \\
Single Branch with Both Tokens (\textit{MobileNetV3}) &   71.4   &  49.2  \\
Ours-Small (w/o AMASS)  &   71.1    & 45.8   \\
Ours-Small  &   68.2    & 44.2   \\
\bottomrule
\end{tabular}
}
}
\caption{
Ablation study on fusion methods, evaluated on the Human3.6M dataset. 
As the Concat Fusion and Add Fusion do not enjoy the ``modality plug-in" characteristics as XFormer, they cannot make use of the MoCap dataset. 
Single-branch model with both tokens does perform better than individual branches but is still worse than two-branch model with cross-modal attention.
For better understanding the contribution of our cross-modal attention mechanism, we also put the results trained without the AMASS dataset. 
We observe that even without AMASS data, our cross-modal attention outperforms Add and Concat Fusion. Concat Fusion even performs worse than individual branch without fusion (\ie, Image Branch Only and Keypoint Branch Only).}
\label{tbl:supp_ablation_fusion}
\end{table}

\mypara{On Fusion Method.}
To further show the effectiveness of our cross-modal attention module, we replace the attention module with other fusion alternatives, \ie, Addition and Concatenation. 
In addition to these two-branches models, we test several single-branch models, including the model containing only an image branch (Image Branch Only), the model containing only a keypoint branch (Keypoint Branch Only), and the model which takes image and keypoint tokens sequentially as input (Single branch with both tokens). The results are shown in Table \ref{tbl:supp_ablation_fusion}. 

We can see that Ours-Small surpasses these naive fusion strategies due to the following reasons:
1) With an attention mechanism, our cross-modal attention module is more effective when modeling relationships of two modalities than simply concatenation or addition. 
2) Addition and Concatenation are more prone to the errors of both branches. As a result, the Concatenation fusion even performs worse than the individual branches.
3) These fusion methods do not have the ``modality plug-in'' characteristic and cannot leverage the MoCap datasets during training.

Compared to the single-branch model, the improvement of the XFormer blocks is more notable for small backbones, as discussed in Section 5.2 in the main paper.

\mypara{On Cross-Modal Attention.}
To verify the efficacy of the cross-modal attention, we further conduct an ablation study as shown in Table \ref{tbl:supp_ablation_cma}. 
For the model w/o Cross-Modal Attention, we remove the cross-modal attention while keeping the self-attention. Further, as our single-branch model has fewer numbers of parameters compared with our full model, we double the filter channels for the single-branch models (denoted as Image Branch $\times 2$ and Keypoint Branch $\times 2$) to show that the performance gain mainly comes from the cross-modal attention rather than the extra parameters. Our full model outperforms these ablations, demonstrating the effectiveness of the cross-modal attention module.

\begin{table}[h]
\small
\renewcommand{\tabcolsep}{1pt}
\centering
{
\begin{tabular}{lcc}
\toprule
Methods         & ~MPJPE~ &  ~PA-MPJPE~ \\ 
\midrule
Image Branch Only (\textit{MobileNetV3})     &   74.5      & 51.3     \\
Keypoint Branch Only (\textit{MobileNetV3})   &   72.8      & 50.5     \\  
w/o Cross-Modal Attention  (\textit{MobileNetV3}) &   69.8   &  48.4  \\
Image Branch $\times 2$  (\textit{MobileNetV3})      &   71.3   &  49.0  \\
Keypoint Branch $\times 2$  (\textit{MobileNetV3})    &   70.2   &  48.3  \\
Ours-Small (w/o AMASS)  &   71.1    & 45.8   \\
Ours-Small  &   68.2    & 44.2   \\
\bottomrule
\end{tabular}
}
\caption{
Ablation study on different settings.
}
\label{tbl:supp_ablation_cma}
\end{table}

\begin{figure*}[t]
 \centering
  \includegraphics[width=0.9\linewidth]{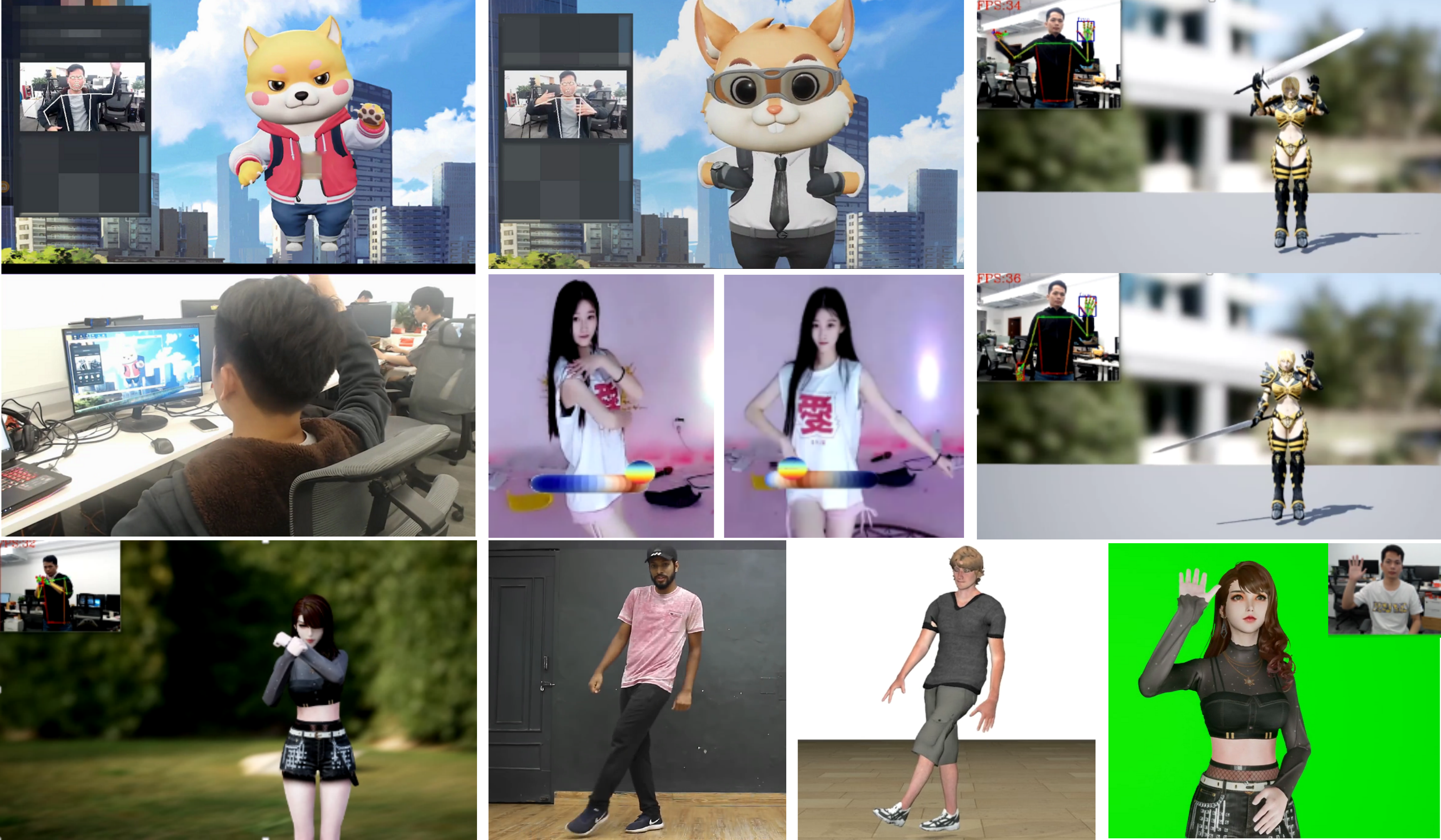}
  \caption{XFormer can be used for virtual character animation and augmented reality effects.}
  \label{fig:app}
\end{figure*}

\mypara{On Consistency Loss and MLP in Cross-Modal Attention.} 
To further demonstrate the effectiveness of consistency loss, we also show the performance of the output of each branch from Ours-Small in Table \ref{tbl:supp_ablation_consistency}.
With the consistency loss, the performance of each branch improves. The improvement of the keypoint branch is relatively higher than that of the image branch.
The consistency loss reduces the gap between the two modalities features in the keypoint branch, resulting in improved full network performance.

Note that w/o consistency loss in Table 3 in main paper still keeps the MLP. 
And when turning off the consistency loss, w/ and w/o MLP have similar performance, as shown in Table \ref{tbl:supp_ablation_consis_mlp}, so updating the features with MLP does not significantly impact the accuracy (slightly higher MPJPE and lower PA-MPJPE).

\begin{table}[h]
\small
\renewcommand{\tabcolsep}{1pt}
\centering
{
\resizebox{\columnwidth}{!}{
\begin{tabular}{lcc}
\toprule
Methods         & ~MPJPE~ &  ~PA-MPJPE~ \\ 
\midrule
w/ MLP (w/ AMASS, w/o  Consistency Loss)    &   70.3   &  45.1  \\
w/o MLP (w/ AMASS, w/o  Consistency Loss)     &   69.6   &  45.5  \\
w/o MLP (w/o AMASS, w/o  Consistency Loss)    &   70.9   &  45.8  \\
Ours-Small  &   68.2    & 44.2   \\
\bottomrule
\end{tabular}
}
}
\caption{
Ablation study on MLP in cross-modal attention module.
}
\label{tbl:supp_ablation_consis_mlp}
\end{table}

\mypara{On Backbones and Different Numbers of XFormer Blocks.} 
In Table \ref{tbl:supp_ablation_settings}, we show the results of different architecture variants of XFormer. We observe that more XFormer blocks lead to higher accuracy, and HRNet yields better performance than ResNet50.

\begin{table}[h]
\centering
\renewcommand{\tabcolsep}{5pt}
\small
{
\begin{tabular}{lcccccc}
\toprule
\multirow{2}{*}{Backbone} & \multicolumn{4}{c}{Number of Blocks} && \multirow{2}{*}{PA-MPJPE} \\
\cline{2-5}
 & $N_{f}$ &$N_{c}$ & $N_{b}$&$N_{x}$ &&  \\ 
\midrule
MobileNetV3 & 0 & 1 & 0 & 1 &&   44.2     \\
MobileNetV3 & 1 & 1 & 1 & 1 &&   43.8     \\
ResNet50 & 1 & 1 & 2 & 2 &&   39.2     \\
HRNet & 1 & 1 & 2 & 2 &&  35.9     \\
HRNet & 1 & 1 & 2 & 3 &&  35.2     \\
\bottomrule
\end{tabular}
}
\caption{
Performance comparison of XFormers with different architectures, evaluated on the Human3.6M dataset.
}
\label{tbl:supp_ablation_settings}
\end{table}

\mypara{Predicting SMPL Shape and Pose Parameters.} 
We also try to use SMPL parameters (shape parameters $\beta \in \mathbb{R} ^ {10}$ and pose parameters $\theta \in \mathbb{R} ^ {24\times3}$) as intermediate representations instead of mesh vertices and joints in our framework and then predict these parameters for 3D body capture. The results are reported in Table \ref{tbl:supp_ablation_poseshape}. Learning with shape and pose parameters is unable to capture the interactions of body mesh and joints, which is incompatible for our framework and thus results in worse reconstruction. 

\begin{table}[h]
\small
\renewcommand{\tabcolsep}{1.5pt}
\centering
{
\begin{tabular}{lcc}
\toprule
Methods         & ~MPJPE~ &  ~PA-MPJPE~  \\ 
\midrule
Ours-Small (Shape and Pose Parameters)              &   73.5       & 47.3     \\ 
Ours-Small (Mesh and Joints) &   68.2    & 44.2   \\
\bottomrule
\end{tabular}
}
\caption{
Ablation study on the types of output, evaluated on the Human3.6M dataset. Modeling interactions of mesh vertices and joints gives higher accuracy than utilizing SMPL shape and pose parameters.
}
\label{tbl:supp_ablation_poseshape}
\end{table}

\section{Discussions}

\mypara{Relation to Minimal-hand \cite{zhou2020monocular}}. 
Although approaching a different task (shape and motion capture of hands), minimal-hand \cite{zhou2020monocular} is related to our work as it aims to use training data with different annotation types. Minimal-hand first utilizes a detector trained on image with 2D labels to predict 2D hand joints. Then, a 3D joint detector trained on 3D annotated images is employed. It further designs an IKNet to map the 3D joint locations to joint rotations, which can be trained on 3D annotated images as well as stand-alone MoCap data. However, this 2D$\rightarrow$3D$\rightarrow$rotations pipeline is built in a sequential fashion, which would cause error accumulation. For example, overfitting in the former module would affect the distribution of input of the latter module, thereby hurting the performance of the end-to-end pipeline. Also, such sequential model prevents information exchanging across different modalities. Instead, our multi-modal-based approach models different modalities in parallel branches and their interactions are explicitly captured with a cross-transformer module. Our method can be similarly applied to capture hand mesh and joints, and we show some 3D hand capture results in the supplementary video.

\mypara{Cross-modal Attention}. Cross-modal attention itself is not new and has been well investigated by recent research \cite{lu2019vilbert,chen2021cascade,wei2020multi,ye2019cross}. XFormer differs by introducing a novel design with the modality switch mechanism and making it work on 3D body capture while existing cross-modal attention work has to train when both modalities are present. Plus, XFormer uses a shared backbone for image and keypoint branch, which exhibits both self- and cross-modal attention characteristics and ensures fast inference. We observe that, in the context of 3D body reconstruction, such cross-modal attention with modality switch is very effective for small backbones while other state-of-the-art methods \cite{lin2021-mesh-graphormer,kocabas2021pare} suffer from performance degradation when the model capacity decreases.

\mypara{Applications.} 
Our method runs at over 30 fps on CPU, so that we can easily apply it to interactive applications. 
As shown in Figure \ref{fig:app}, our method can produce accurate and temporal coherent character animation results in commercial engines. 
Further, our framework captures accurate human motion both in whole body and upper body scenes as shown in Figure \ref{fig:app}. This makes our system perfectly suitable for Virtual YouTubers, as they usually live steam with only their upper body shown in the camera with hip stay still.
Moreover, since our method reconstructs the mesh of a person, we can enable a set of AR effects. For example, as shown in the middle of Figure \ref{fig:app}, the ``Body Stickers'' puts a virtual sticker on to a person's body according to the predicted mesh and tracks the body's movement in real-time. Please find more video results of these applications in the supplementary video.

\end{document}